\documentclass[10pt,conference,final]{IEEEtran}
\IEEEoverridecommandlockouts
\usepackage{cite}
\usepackage{amsmath,amssymb,amsfonts}
\usepackage{algorithmic}
\usepackage{graphicx}
\usepackage{textcomp}
\usepackage{xcolor}
\usepackage{booktabs}   
\usepackage{multirow}   
\usepackage[table]{xcolor} 
\usepackage{colortbl}   

\usepackage{array}      
\usepackage{caption}    
\usepackage{subcaption} 
\usepackage[hidelinks]{hyperref}
\def\BibTeX{{\rm B\kern-.05em{\sc i\kern-.025em b}\kern-.08em
    T\kern-.1667em\lower.7ex\hbox{E}\kern-.125emX}}
\begin{document}

\title{MotiMem: Motion-Aware Approximate Memory for Energy-Efficient Neural Perception in Autonomous Vehicles
}

\author{%
	\IEEEauthorblockN{Haohua Que\textsuperscript{1,\textdagger}, Mingkai Liu\textsuperscript{2,\textdagger}, Jiayue Xie\textsuperscript{3}, Haojia Gao\textsuperscript{3},
		Jiajun Sun\textsuperscript{3}, Hongyi Xu\textsuperscript{3,4}, Handong Yao\textsuperscript{1,*}, Fei Qiao\textsuperscript{3,*}}%
}
\maketitle
\begingroup
\renewcommand\thefootnote{}
\footnotetext{\footnotesize

	\textsuperscript{1}University of Georgia;
	\textsuperscript{2}Peking University;
	\textsuperscript{3}Tsinghua University;
	\textsuperscript{4}Infinity Robotics;
  \textsuperscript{\textdagger}Co-first authors: Haohua Que and Mingkai Liu.
  \textsuperscript{*}Corresponding authors: Handong Yao (Handong.Yao@uga.edu) and Fei Qiao (qiaofei@tsinghua.edu.cn).
	This work was supported by the Beijing Natural Science Foundation (L253009);
	the National Natural Science Foundation of China (62334006, U25A20489);
	the Brain Science and Brain-like Intelligence Technology National Science and Technology Major Project of China (Grant No.~2025ZD0215600);
	and the National Key Technologies R\&D Program of China (2025YFF1500600).
	Haohua Que and Handong Yao completed their work entirely in the United States and received no funding support.
}
\endgroup
\setcounter{footnote}{0}

\begin{abstract}
	High-resolution sensors are critical for robust autonomous perception but impose a severe ``memory wall" on battery-constrained electric vehicles. In these systems, data movement energy often outweighs computation. Traditional image compression is ill-suited as it is semantically blind and optimizes for storage rather than bus switching activity.
	We propose \textbf{MotiMem}, a hardware-software co-designed interface. Exploiting temporal coherence, MotiMem uses lightweight 2D Motion Propagation to dynamically identify Regions of Interest (RoI). Complementing this, a Hybrid Sparsity-Aware Coding scheme leverages adaptive inversion and truncation to induce bit-level sparsity.
	Extensive experiments across nuScenes, Waymo, and KITTI with 16 detection models demonstrate that MotiMem reduces memory-interface dynamic energy by \textbf{$\approx 43\%$} while retaining \textbf{$\approx 93\%$} of the object detection accuracy, establishing a new Pareto frontier significantly superior to standard codecs like JPEG and WebP.
\end{abstract}


\section{Introduction}

\begin{figure}[t] 
	\centering

	\includegraphics[width=\linewidth]{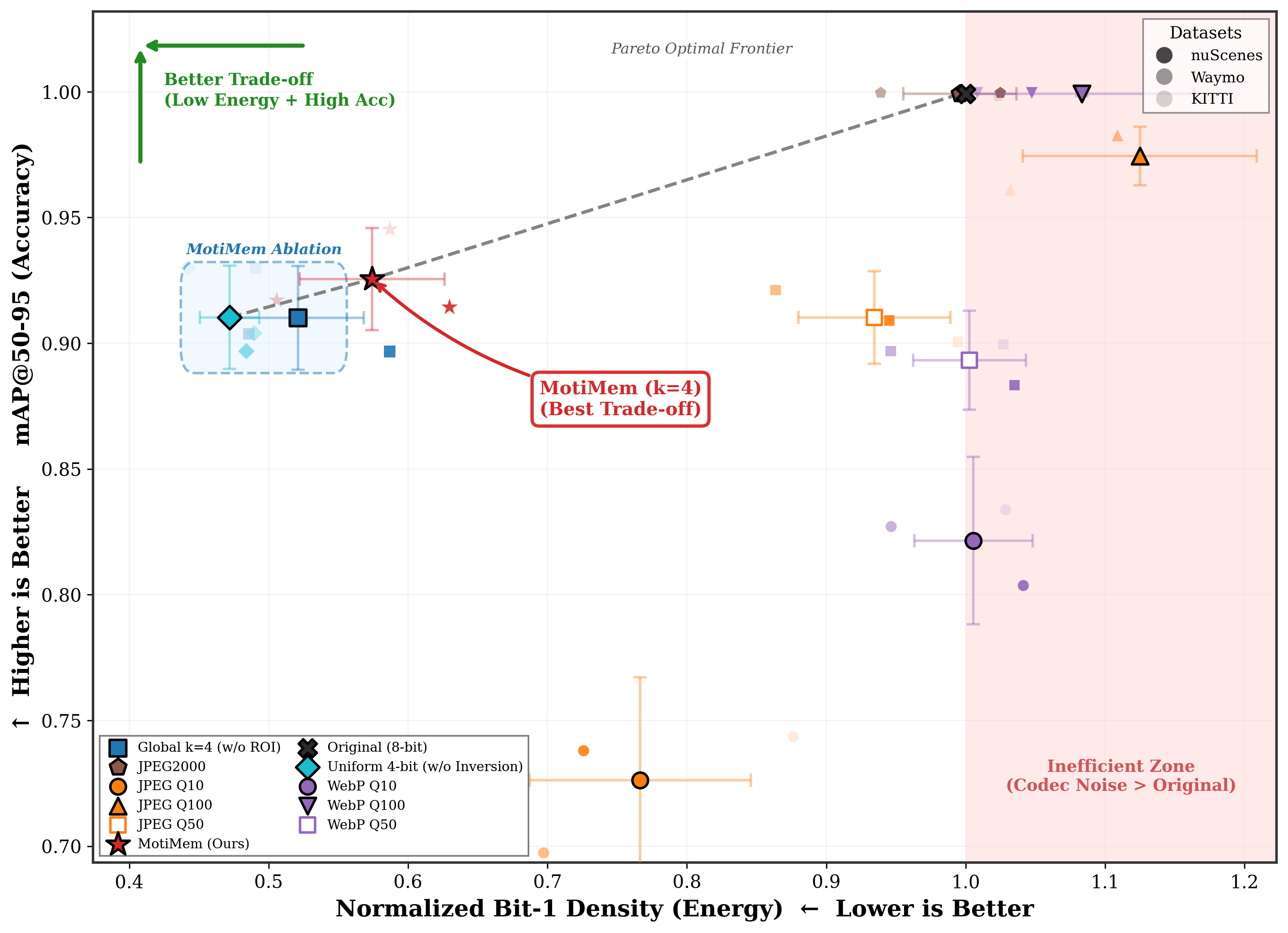}

	\caption{\textbf{Pareto Efficiency Analysis (Accuracy vs. Energy Proxy) across nuScenes, Waymo, and KITTI Datasets~\cite{caesar2020nuscenes,sun2020scalability,geiger2012we,geiger2013vision}.}
		MotiMem (marked with a \textcolor{red}{red star}) establishes a superior Pareto frontier compared to standard codecs (JPEG, WebP) which fall into the inefficient zone due to high entropy.
		Evaluated across \textbf{three diverse datasets} and \textbf{16 detection models}, MotiMem achieves the optimal trade-off by reducing the \textbf{normalized bit-1 density} (a direct proxy for dynamic energy) by $\approx$ 43\% while maintaining $\approx$ 93\% of the original uncompressed mAP.
		Crucially, MotiMem significantly outperforms the semantically blind ``Global k=4'' ablation (blue square), validating that \textbf{semantic-aware allocation} is superior to uniform quantization for energy-constrained perception.}
	\label{fig:pareto_efficiency}
	\vspace{-19pt}
\end{figure}


In the rapidly evolving field of autonomous vehicles (AVs) and mobile robotics,
visual perception serves as the cornerstone of safety and decision-making.
To detect distant obstacles and ensure reliable navigation, modern perception
systems of AVs are increasingly equipped with high-resolution cameras (e.g., 4K or 8K)
\cite{baris2025automotive,yao2023radar}.
However, this pursuit of visual fidelity imposes a severe penalty on the underlying
hardware. The massive influx of high-dimensional visual data creates a significant
bottleneck in memory bandwidth and system energy consumption \cite{liu2019edge,becker2020demystifying}.
For battery-constrained edge devices, the energy cost of data movement and storage
often exceeds that of computation itself \cite{kestor2013quantifying,pandiyan2014quantifying},
exacerbating the well-known ``Memory Wall'' problem \cite{saulsbury1996missing,zhang2023survey}.
Consequently, designing energy-efficient perception systems without compromising
detection accuracy has become a critical challenge \cite{guo2018embedded}.


To mitigate this memory pressure, conventional approaches typically resort to
standard image compression standards (e.g., JPEG, WebP).
While effective in reducing data volume, these methods are ill-suited for the
``sensing-for-perception'' pipeline for two fundamental reasons.
First, they are \textit{semantically blind}: they treat irrelevant background
pixels (e.g., sky, road surface) with the same importance as safety-critical
foreground objects, wasting precious bandwidth on non-informative content
\cite{wang2023semantic,lohdefink2020focussing}.
Second, and more critically from a hardware perspective, standard codecs optimize
solely for storage capacity rather than \textbf{bit-level activity}
\cite{stan2002bus,ramprasad2002coding}.
The resulting compressed bitstreams often exhibit high entropy (randomness), which maximizes the switching activity factor ($\alpha$) across \textbf{both off-chip buses and RAM}. \cite{petrov2004low,lekatsas2005approximate}.
This leads to excessive dynamic power dissipation during data transfer,
negating the energy benefits of reduced volume \cite{panda2002low,benini1997asymptotic}.
Furthermore, complex decoding processes introduce additional latency, which is
detrimental to real-time control loops \cite{faykus2022lossy,pawlowski2024lossy}.


To bridge the gap between high-fidelity sensors and energy-efficient neural
networks, we propose \textbf{MotiMem}, a hardware-software co-designed neural
interface \cite{ponzina2022hardware,karageorgos2020hardware}.
Unlike passive data pipes, MotiMem actively reshapes the sensor stream based
on the \textit{temporal coherence} inherent in driving scenarios
\cite{leon2021review,wu2020motionnet}.
Our key insight is that physical objects do not move randomly; their continuous
motion allows us to utilize detection results from the previous frame ($T-1$)
as a low-cost motion prior to predict Regions of Interest (RoI) for the
current frame ($T$) \cite{xie2025cohere3d}.
Complementing this kinematic awareness, we introduce a
\textbf{\textit{Hybrid Sparsity-Aware Coding}} scheme.
To maximize bus and memory efficiency, this approach differentially encodes
data based on semantic importance: for critical RoIs, it applies
\textbf{high-bit inversion} with a Least Significant Bit (LSB) flag to
maintain full precision; for the background, it couples inversion with
aggressive \textbf{low-bit truncation} \cite{doukas2007region,liu2024roi}.
This hybrid strategy induces high bit-level sparsity
\cite{song2015more}, which directly minimizes switching
activity across both the sensor-memory bus and the RAM arrays themselves
\cite{stan2002bus,ramprasad2002coding}.
This allows the system to preserve target fidelity while significantly reducing memory-interface energy for robust neural-based autonomous perception
(e.g., object detection).
Recent advances in other high-stakes learning domains also reinforce this design intuition: structure-preserving graph modeling and multi-scale temporal fusion improve predictive robustness when task-relevant spatial/temporal cues are retained\cite{zhang2025spsgnn,liu2026mstdp}.

Our main contributions are:
\begin{itemize}
	\item We propose \textbf{MotiMem}, a hardware--software co-designed neural interface that leverages temporal coherence for non-uniform precision storage. It identifies RoIs via kinematic-guided motion propagation, avoiding the latency of heavy motion estimation.
	\item We introduce a \textbf{\textit{Hybrid Sparsity-Aware Coding}} scheme that couples adaptive inversion with background truncation. It reduces the \textbf{normalized bitstream activity} (proxy for memory-interface dynamic energy) by $\approx$ 43\% (Fig.~\ref{fig:pareto_efficiency}) while preserving task-critical fidelity.
	\item We conduct extensive experiments on video object detection across \textbf{nuScenes, Waymo, and KITTI}~\cite{caesar2020nuscenes,sun2020scalability,geiger2012we,geiger2013vision} and \textbf{16 detection models}, demonstrating a superior accuracy--energy Pareto frontier (Fig.~\ref{fig:pareto_efficiency}).
\end{itemize}

\section{Related Work}

\subsection{Perception Stream Compression}
Standard codecs (e.g., JPEG, WebP) are primarily optimized for \textit{Human Visual System (HVS)} fidelity (e.g., PSNR/SSIM) rather than downstream machine vision accuracy~\cite{dong2015compression, hao2023understanding}.
They rely on block-based prediction/transform coding with quantization that suppresses high-frequency details, which can distort edge and texture cues important to CNNs~\cite{dong2015compression, dodge2016understanding}.
Critically, such approaches typically disregard the \textbf{hardware energy cost} of data movement~\cite{horowitz2014computing,chen2016eyeriss}.
Moreover, many hardware-friendly compression/quantization schemes are \textit{semantically blind}~\cite{wang2019haq, gholami2021survey}, allocating uniform precision to both irrelevant background (e.g., sky) and safety-critical foreground objects.
While learning-based codecs (e.g., VAE-based) can achieve superior rate--distortion performance~\cite{balle2018hyperprior,minnen2018joint,toderici2017fullres}, their neural decoding often incurs \textbf{prohibitive latency and compute/energy overhead} on edge hardware~\cite{sze2017dnn_survey,lu2019dvc}.
Therefore, we compare against industry-standard codecs and hardware-friendly quantization baselines that represent the current low-latency operating point in automotive perception~\cite{wiegand2003h264,sullivan2012hevc,iso10918jpeg,jacob2018quant}.
In contrast, \textbf{MotiMem} introduces a power-centric, RoI-guided coding paradigm: rather than only reducing bandwidth, it reduces the density of bit-1s and the \textbf{effective transition activity} in the storage stream, inducing high sparsity without necessarily reducing raw interface bandwidth.
This directly lowers the \textbf{(weighted) switching activity} and thus dynamic power dissipation in RAM and bus interfaces~\cite{chandrakasan1992lowpower,stan1997businvert}.

\subsection{Perception-Aware RoI Prioritization}
Different from codec-centric compression, these works prioritize \emph{where} to preserve information for perception tasks.
Recent studies on perception-aware visual systems investigate how to prioritize task-relevant regions to improve downstream robustness under constrained resources.
Such methods range from semantic/object-centric region selection to RoI-driven quality allocation and foveated processing, especially in autonomous driving and networked perception settings~\cite{wang2023semantic,lohdefink2020focussing,doukas2007region,liu2024roi}.
Nevertheless, two limitations hinder their deployment in real-time AV pipelines.
First, RoI estimation is often coupled with additional computation (e.g., motion estimation or multi-stage semantic modules), which increases latency and complicates system integration~\cite{faykus2022lossy,pawlowski2024lossy}.
Second, these approaches typically optimize rate--accuracy (or rate--distortion) objectives, without explicitly targeting \emph{memory-interface energy} metrics such as switching activity during storage and transfer~\cite{horowitz2014computing,chen2016eyeriss,chandrakasan1992lowpower,stan1997businvert}.
MotiMem addresses these limitations by deriving RoIs from prior detections via lightweight motion propagation and by shaping the stored bitstream to reduce memory-interface dynamic power.

\subsection{Energy-Aware Neural Interfaces and Approximate Memory}
Energy-efficient neural perception increasingly recognizes that the system-level energy is often dominated by memory access and data movement rather than arithmetic~\cite{horowitz2014computing,chen2016eyeriss}.
At the interface level, dynamic power on buses and memory I/O is strongly governed by switching activity and data-dependent bit transitions, motivating transition-aware encoding schemes~\cite{chandrakasan1992lowpower,stan1997businvert}.
Recent low-power memory designs further exploit bit-level manipulation, e.g., selective bit dropping combined with encoding co-strategies for DRAM/SRAM, to reduce switching and I/O energy~\cite{10006792,liu2025new,jiang2015approximate, liu2019approximate}.
In parallel, approximate computing and low-precision inference reduce storage/compute bit-width with controlled accuracy loss~\cite{jacob2018quant,gholami2021survey,wang2019haq}.
However, existing methods are often task-agnostic or semantically uniform, applying approximation/encoding without accounting for which regions are critical to neural perception.
MotiMem bridges this gap by co-designing motion-guided RoI semantics with sparsity-aware bitstream shaping (inversion and background truncation), directly targeting memory-interface dynamic power while preserving task-critical accuracy.
\begin{figure*}[th]
	\centering
	\includegraphics[width=\linewidth]{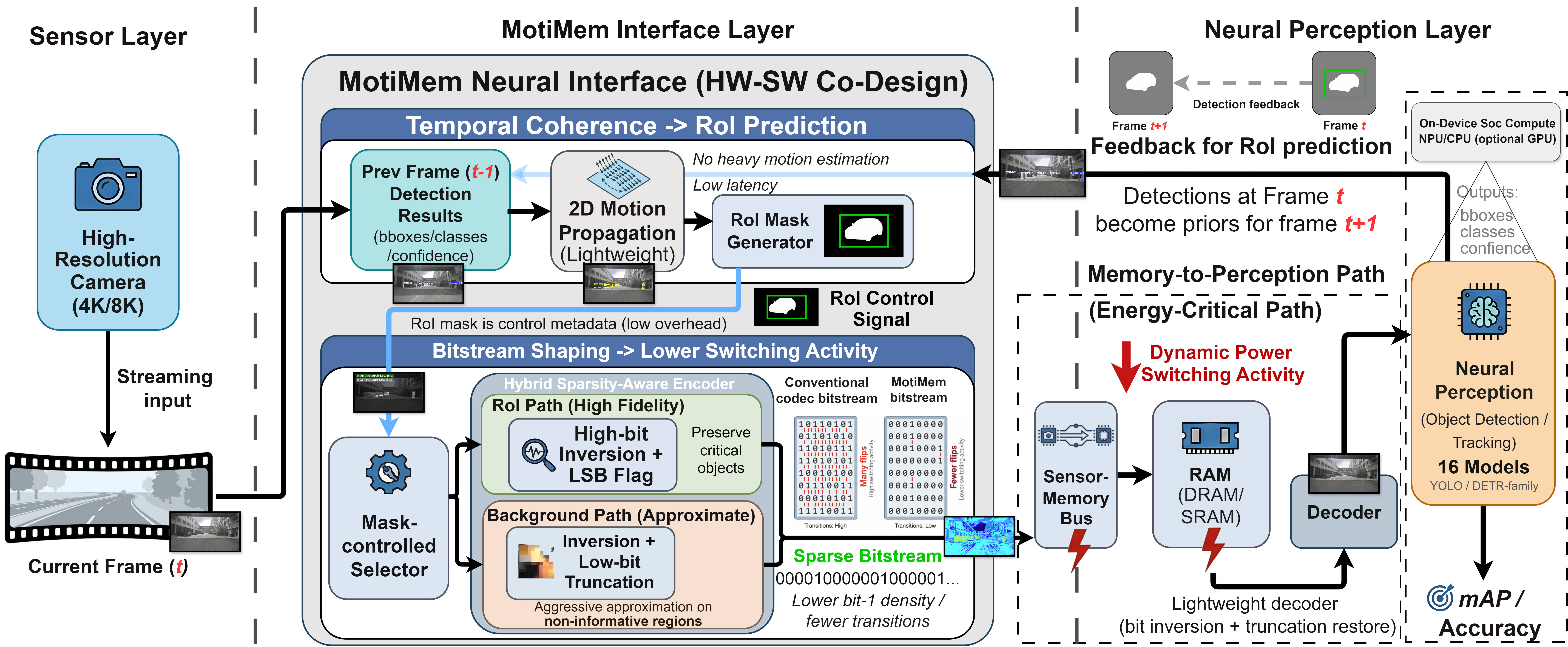}
	\caption{\textbf{The MotiMem closed-loop interface for neural perception.}
		MotiMem sits between the camera stream and the memory hierarchy.
		It forms a closed loop: detections at time $t$ are fed back as compact metadata to predict the RoI mask for time $t{+}1$.
		Guided by the RoI mask, MotiMem applies RoI-guided hybrid coding to shape the stored bitstream toward lower activity (fewer 1s/toggles) along the sensor-to-memory path, while preserving perception accuracy.}
	\label{fig:overview}
	\vspace{-10pt}
\end{figure*}

\section{Methodology}
\label{sec:methodology}

\subsection{Problem Setup and System Goal}
We study a streaming on-device perception pipeline for autonomous driving.
At each time step $t$, a high-resolution camera frame
$ I_t \in \mathbb{R}^{H \times W \times C} $
is captured and stored in memory before being consumed by a detector/tracker running on an on-device compute unit (CPU/NPU).
The memory path (sensor $\rightarrow$ bus $\rightarrow$ RAM) is often energy-critical due to data-dependent activity on buses and memory writes.
Unlike codec-centric designs that optimize only storage size, MotiMem targets \textbf{system-level energy} by shaping the \emph{bit statistics} of the representation stored/transferred along the memory path, while maintaining detection/tracking quality.
MotiMem takes the current frame $I_t$ and the previous detection results $D_{t-1}$ (bounding boxes, classes, confidence scores) as input. It outputs an encoded stream stored in RAM, decoded with lightweight bit operations, and fed into standard perception models.

\subsection{Closed-Loop Overview}
\label{sec:overview}
MotiMem consists of two coupled components (Fig.~\ref{fig:overview}):
(1) \textbf{Temporal Coherence $\rightarrow$ RoI Prediction:} leverage temporal coherence to predict a binary RoI mask $M_t$ for the current frame using the prior detections $D_{t-1}$.
(2) \textbf{RoI-Guided Bitstream Shaping:} use $M_t$ to apply a hybrid coding strategy where RoI is coded with high fidelity (but not strictly lossless), and the background is aggressively simplified. Both paths apply bit-level shaping to bias the stored stream toward fewer 1s and fewer toggles. The key system intuition is that only a small portion of each frame is critical for perception, and thus representation fidelity can be allocated non-uniformly.

\subsection{Temporal Coherence and RoI Prediction}
\label{sec:roi_prediction}

\begin{figure}[t]
	\centering
	\includegraphics[width=\linewidth]{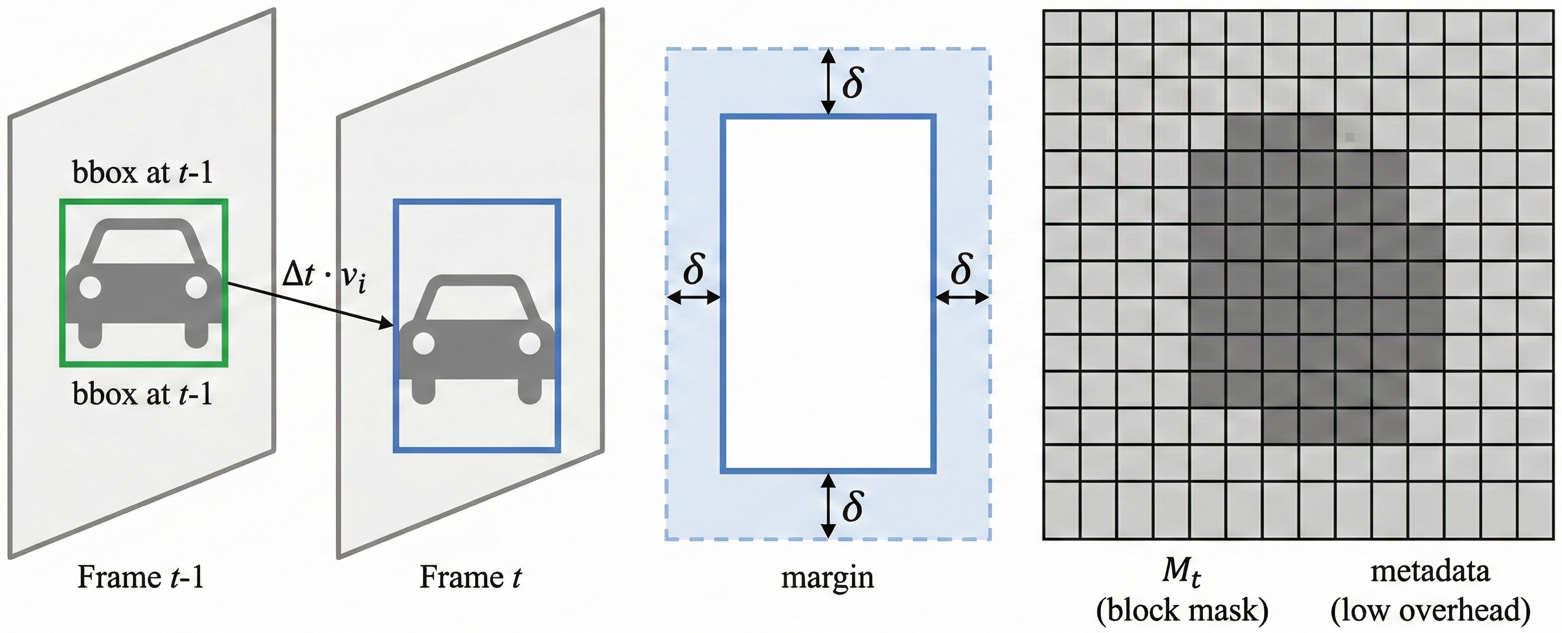}
	\caption{\textbf{RoI prediction from temporal coherence.}
		Detections at frame $t{-}1$ are propagated to frame $t$ using a lightweight 2D motion prior, inflated by margin $\delta$, and rasterized to a compact block-level RoI mask $M_t$ (e.g., $16{\times}16$ blocks).}
	\label{fig:ROI}
	\vspace{-10pt}
\end{figure}
As shown in Fig.~\ref{fig:ROI}, MotiMem predicts the RoI mask $M_t$ for the current frame $I_t$ by propagating prior detections $D_{t-1}$ using lightweight 2D motion estimation.
Let the detector output at the previous frame be
\begin{equation}
	D_{t-1} = \{(b_i, c_i, \gamma_i)\}_{i=1}^{N_{t-1}},
\end{equation}
where $b_i$ is a bounding box, $c_i$ is class, and $\gamma_i$ is confidence.

\paragraph{Lightweight 2D motion propagation.}
Instead of optical flow, we propagate each box in the image plane using a constant-velocity model:
\begin{equation}
	\hat{b}_i^{(t)} = b_i^{(t-1)} + \Delta t \cdot \mathbf{v}_i,
	\label{eq:box_prop_sys}
\end{equation}
where $\mathbf{v}_i$ is a lightweight 2D velocity estimate (e.g., from recent box center displacement or a simple tracker).
To tolerate uncertainty, we inflate the predicted box by margin $\delta$:
\begin{equation}
	\tilde{b}_i^{(t)} = \mathrm{Inflate}(\hat{b}_i^{(t)}, \delta).
\end{equation}

\paragraph{RoI mask as control metadata (low overhead).}
We build a binary RoI mask by the union of inflated boxes:
\begin{equation}
	M_t(p)=\mathbb{I}\Big[p \in \bigcup_{i=1}^{N_{t-1}} \tilde{b}_i^{(t)}\Big],
	\quad p\in\{1,\ldots,H\}\times\{1,\ldots,W\}.
	\label{eq:mask_sys}
\end{equation}
In practice, $M_t$ is represented at block granularity (e.g., $16{\times}16$ blocks), so it is compact control metadata that does \emph{not} traverse the high-volume pixel stream and introduces negligible overhead relative to $I_t$.

\subsection{RoI-Guided Hybrid Coding with a Single Parameter $k$}
\label{sec:hybrid_coding}

\begin{figure}[h]
	\centering
	\includegraphics[width=\linewidth]{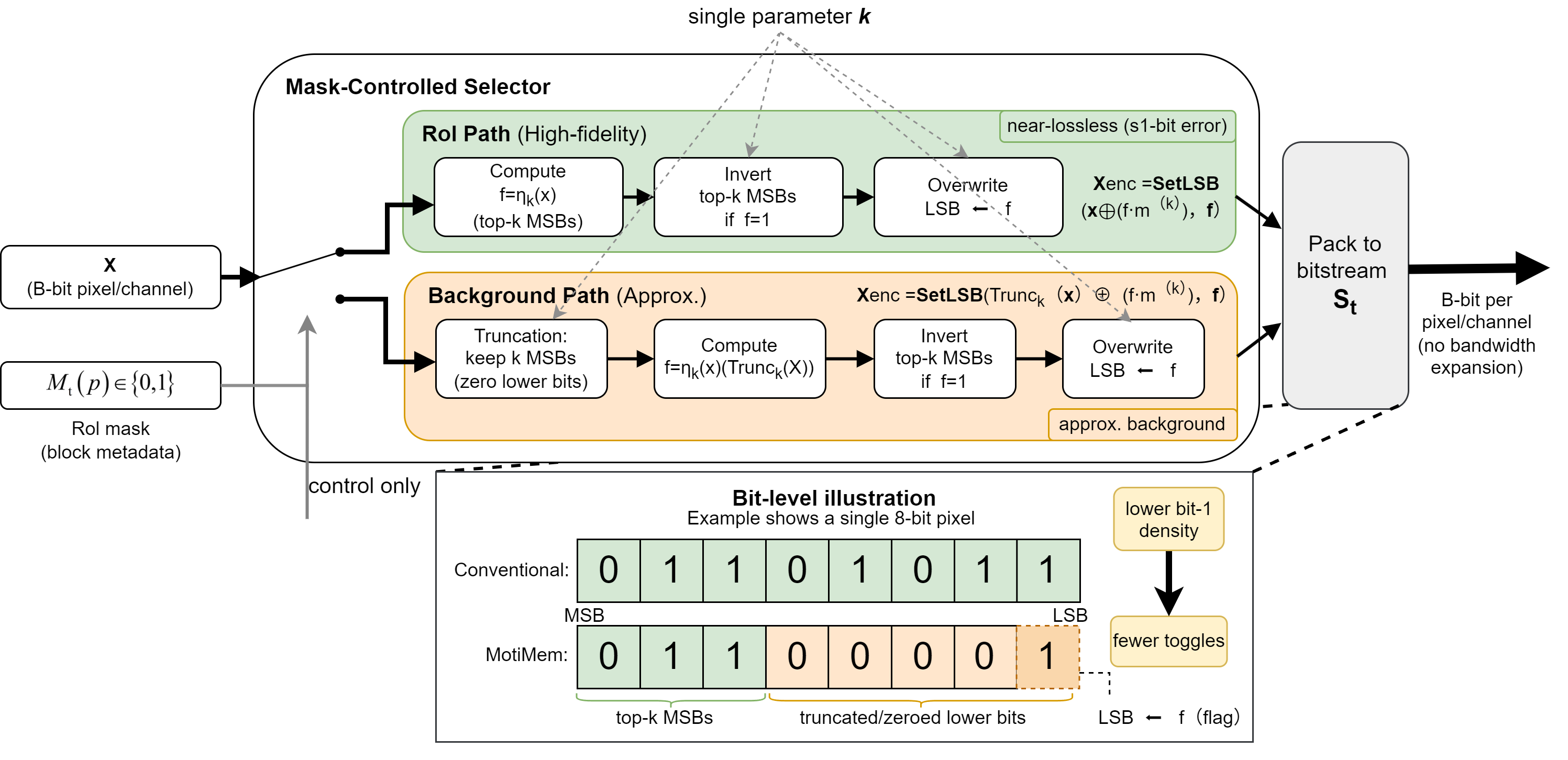}
	\caption{\textbf{RoI-guided hybrid coding.}
		Pixels are routed into two paths based on $M_t(p)$ and parameter $k$.
		\textbf{RoI:} Compute and embed an inversion flag $f$ in the LSB, selectively invert top-$k$ MSBs if $f{=}1$.
		\textbf{Background:} Truncate to top-$k$ MSBs, apply the same inversion, and embed $f$ in the LSB.
		Both paths preserve $B$-bit width, reducing bit-1 density and transitions for lower memory energy.}

	\label{fig:HybridCode}
	\vspace{-10pt}
\end{figure}

We encode the current frame $I_t$ into a shaped representation by routing pixels into two paths controlled by $M_t$, as shown in Fig.~\ref{fig:HybridCode}.
Our design is inspired by low-power bit dropping/encoding strategies for memory interfaces~\cite{10006792,liu2025new},
and extends them to \emph{neural-perception-aware} coding via RoI-guided non-uniform fidelity.

We use a single integer parameter $k$ to control both:
(i) background approximation strength by retaining only the top-$k$ MSBs, and
(ii) RoI bit shaping by selectively inverting the top-$k$ MSBs, while using the \emph{pixel LSB as an embedded inversion flag}.
This choice keeps the stored pixel width unchanged ($B$ bits per pixel/channel), avoiding raw interface bandwidth expansion.

Let $x_{t,p}\in\{0,\dots,2^B-1\}$ denote a $B$-bit pixel/channel value at location $p$.
We decompose $x$ into MSBs and the least-significant bit (LSB):
\begin{equation}
	x = 2 \cdot \bar{x} + \ell(x),
\end{equation}
where $\ell(x)\in\{0,1\}$ is the LSB and $\bar{x}\in\{0,\dots,2^{B-1}-1\}$ contains the upper $(B-1)$ bits.

Define the top-$k$ MSB mask over the full $B$-bit word:
\begin{equation}
	\mathbf{m}^{(k)} \triangleq \sum_{j=B-k}^{B-1} 2^j.
\end{equation}

\subsubsection{RoI path: high-fidelity selective inversion with LSB-embedded flag}
For RoI pixels ($M_t(p)=1$), we preserve perceptual fidelity while shaping bit statistics.
We decide whether to invert the top-$k$ MSBs based on their Hamming weight:
\begin{equation}
	\eta_k(x) \triangleq \mathbb{I}\!\left[w_H\!\big(x \wedge \mathbf{m}^{(k)}\big) > \tau\right],
	\qquad \tau \in [0,k],
	\label{eq:eta_sys}
\end{equation}
where we typically use $\tau = k/2$.

\paragraph{Flag embedding.}
We embed the inversion decision into the LSB, i.e., we overwrite the LSB with the flag:
\begin{equation}
	f(p) \triangleq \eta_k(x), \qquad \ell(x^{\text{RoI}}_{\text{enc}}) = f(p).
\end{equation}

\paragraph{Selective inversion of MSBs.}
We invert only the top-$k$ bits when $f(p)=1$:
\begin{equation}
	x^{\text{RoI}}_{\text{enc}}
	= \Big( x \oplus \big(f(p)\cdot \mathbf{m}^{(k)}\big) \Big) \;\;\text{with LSB overwritten by } f(p).
	\label{eq:roi_enc_lsbflag}
\end{equation}
This operation is \emph{near-lossless}: only the original LSB may be lost because it is reused as the embedded flag, while the inverted MSB range is reversible given $f(p)$.
\vspace{-2.5pt}
\paragraph{Lightweight decoding.}
At the decoder, we read the embedded flag from the LSB and invert back the same MSB range:
\begin{equation}
	\hat{f}(p) = \ell(x^{\text{RoI}}_{\text{enc}}),
	\qquad
	\hat{x} = x^{\text{RoI}}_{\text{enc}} \oplus \big(\hat{f}(p)\cdot \mathbf{m}^{(k)}\big).
	\label{eq:roi_dec_lsbflag}
\end{equation}
Because the original LSB was overwritten, $\hat{x}$ matches $x$ up to a 1-bit error:
\begin{equation}
	|x - \hat{x}| \le 1,
\end{equation}
which empirically preserves detection accuracy (Fig.~\ref{fig:pareto_efficiency}).

\subsubsection{Background path: $k$-MSB truncation + MSB inversion with LSB-embedded flag}
For background pixels ($M_t(p)=0$), MotiMem enforces sparsity by a fixed two-step transform: \textbf{truncate} the low bits and \textbf{shape} the retained MSBs via selective inversion. The inversion decision is embedded into the pixel LSB as a flag, so the stream width remains $B$ bits per pixel/channel.
\vspace{-2.5pt}
\paragraph{Truncate low bits.}
We keep only the top-$k$ MSBs and zero the remaining $(B-k)$ bits:
\begin{equation}
	x^{(k)} \triangleq \mathrm{Trunc}_k(x)
	= \left\lfloor \frac{x}{2^{B-k}} \right\rfloor \cdot 2^{B-k}.
	\label{eq:trunc_sys}
\end{equation}

\paragraph{Invert dense MSB patterns and write the flag into LSB.}
We compute an inversion indicator from the truncated value
\begin{equation}
	f(p) \triangleq \eta_k\!\left(x^{(k)}\right)
	= \mathbb{I}\!\left[w_H\!\big(x^{(k)} \wedge \mathbf{m}^{(k)}\big) > \tau\right],
	\label{eq:bg_flag}
\end{equation}
and apply inversion on the same top-$k$ MSB range:
\begin{equation}
	\tilde{x}^{(k)} \triangleq x^{(k)} \oplus \big(f(p)\cdot \mathbf{m}^{(k)}\big).
	\label{eq:bg_invert}
\end{equation}
Finally, we overwrite the pixel LSB with the flag:
\begin{equation}
	x^{\text{BG}}_{\text{enc}} \triangleq \mathrm{SetLSB}\!\big(\tilde{x}^{(k)},\, f(p)\big),
	\qquad \ell\!\left(x^{\text{BG}}_{\text{enc}}\right)=f(p).
	\label{eq:bg_enc_sys}
\end{equation}

This background coding is intentionally approximate due to truncation (and LSB reuse), but it produces a sparse, low-activity stream that reduces normalized bit-1 density and toggle rate along the memory path.

\subsubsection{Mask-controlled routing and bitstream packing}
The per-pixel routing is:
\begin{equation}
	x_{\text{enc}}(p)=
	\begin{cases}
		x^{\text{RoI}}_{\text{enc}}(p), & M_t(p)=1, \\
		x^{\text{BG}}_{\text{enc}}(p),  & M_t(p)=0.
	\end{cases}
	\label{eq:routing_sys}
\end{equation}
The resulting encoded values are stored/transferred as a standard $B$-bit-per-pixel stream $\mathbf{s}_t$ (same tensor shape as $I_t$).
Importantly, the inversion decision is embedded into the pixel LSB in \emph{both} paths, so the representation width is unchanged and no additional raw interface bandwidth is introduced.
The RoI mask $M_t$ is compact control metadata (block-level) used only to steer routing/encoding and does not traverse the high-volume pixel stream. In our implementation, $M_t$ is generated locally at the interface from $D_{t-1}$ and thus does not require transmitting per-pixel side information.

\subsection{Bitstream Activity and Energy Proxy Metrics}
\label{sec:activity_metrics}
MotiMem reduces system energy by lowering the \emph{data-dependent activity} of the stored/transferred bitstream on the sensor--memory path. We report two proxies~\cite{10006792,liu2025new}.

\paragraph{Normalized Bit-1 density.}
For a frame-level bitstream $\mathbf{s}\in\{0,1\}^{L}$,
\begin{equation}
	\rho_1(\mathbf{s}) \triangleq \frac{1}{L}\sum_{\ell=1}^{L} s_\ell,
	\qquad
	\mathrm{NBD}_t \triangleq
	\frac{\rho_1(\mathbf{s}^{\text{enc}}_t)}{\rho_1(\mathbf{s}^{\text{raw}}_t)}.
	\label{eq:nbd_def}
\end{equation}
Smaller $\mathrm{NBD}_t$ means fewer ``1'' bits in the encoded stream (higher sparsity), directly reducing dynamic energy.

\paragraph{Bit-transition activity}
Partition $\mathbf{s}$ into $W$-bit words $\{\mathbf{s}_n\}_{n=1}^{N}$. The normalized transition rate is
\begin{equation}
	\alpha(\mathbf{s}) \triangleq \frac{1}{(N-1)W}\sum_{n=2}^{N} d_H(\mathbf{s}_n,\mathbf{s}_{n-1}),
	\label{eq:toggle_sys}
\end{equation}
where $d_H(\cdot,\cdot)$ is the Hamming distance.

\paragraph{Effect of shaping}
Truncation (Eq.~\eqref{eq:trunc_sys}) forces most low-order bits to zero (except the LSB flag), and selective MSB inversion (Eq.~\eqref{eq:roi_enc_lsbflag}, \eqref{eq:bg_enc_sys}) suppresses dense-one MSB patterns. Together, they reduce $\rho_1(\mathbf{s}^{\text{enc}})$ and $\alpha(\mathbf{s}^{\text{enc}})$, while using only lightweight bitwise operations.

\subsection{Decoding and Closed-Loop Inference}
Finally, the SoC performs lightweight decoding with only bitwise operations.
For both RoI and background pixels, the decoder reads the embedded flag from the pixel LSB and reverses the selective top-$k$ MSB inversion (Eq.~\eqref{eq:roi_dec_lsbflag}).
For background pixels, the lower bits remain truncated (i.e., zeroed except the LSB used as the flag), yielding an approximate reconstruction.
The decoded frame is then processed by standard detectors/trackers (e.g., YOLO/DETR-family), and the resulting detections are fed back as $D_t$ to generate $M_{t+1}$, closing the loop. This feedback forms a closed-loop interface where perception outputs at $t$ act as control metadata for encoding at $t{+}1$.

\vspace{-2.5pt}
\section{Experimental Evaluation}
\label{sec:experiments}

\subsection{Experimental Setup}
\paragraph{Datasets \& Models} \textbf{nuScenes}~\cite{caesar2020nuscenes}, \textbf{Waymo}~\cite{sun2020scalability}, and \textbf{KITTI}~\cite{geiger2012we,geiger2013vision}. To ensure robustness, we report average performance across \textbf{16 diverse object detection models}, including the YOLO family (v5, v8, v9, v10, v11, YOLO26) and Transformer-based RT-DETR.

\paragraph{Baselines} (1) \textbf{Standard Codecs:} Industry standards including \textit{JPEG} and \textit{WebP} (configured at Q10, Q50, Q100), and lossless \textit{JPEG2000}; (2) \textbf{Ablations:} To isolate component contributions, we evaluate \textit{Global $k=4$ (w/o ROI)} (applying uniform truncation to the whole frame) and \textit{Uniform 4-bit (w/o Inversion)} (disabling the adaptive bit-flip optimization); (3) \textbf{Original (8-bit):} The uncompressed raw baseline, serving as the upper bound for accuracy.

\paragraph{Metrics} We report performance on two axes: (1) \textbf{Perception Accuracy}, measured by standard Mean Average Precision (\textbf{mAP@50-95}); and (2) \textbf{Energy Efficiency}, quantified by the \textbf{Normalized Bit-1 Density Ratio}. As defined in Eq.~\ref{eq:nbd_def}, this metric serves as a linear proxy for dynamic switching power in memory interfaces, where a ratio below 1.0 indicates effective energy reduction compared to the uncompressed baseline.

\paragraph{Implementation Details}
The MotiMem encoding logic and detection pipeline were \textbf{functionally simulated} in software using PyTorch.
Critically, $D_{t-1}$ is obtained from \textbf{actual model detections} on the previously encoded frame (not ground-truth), faithfully emulating the closed-loop deployment.
All experiments were conducted on a workstation with an Intel Core Ultra 9 285K CPU, 128GB RAM, and an NVIDIA RTX 6000 Pro GPU.

\begin{figure}[t]
	\centering
	\includegraphics[width=\linewidth]{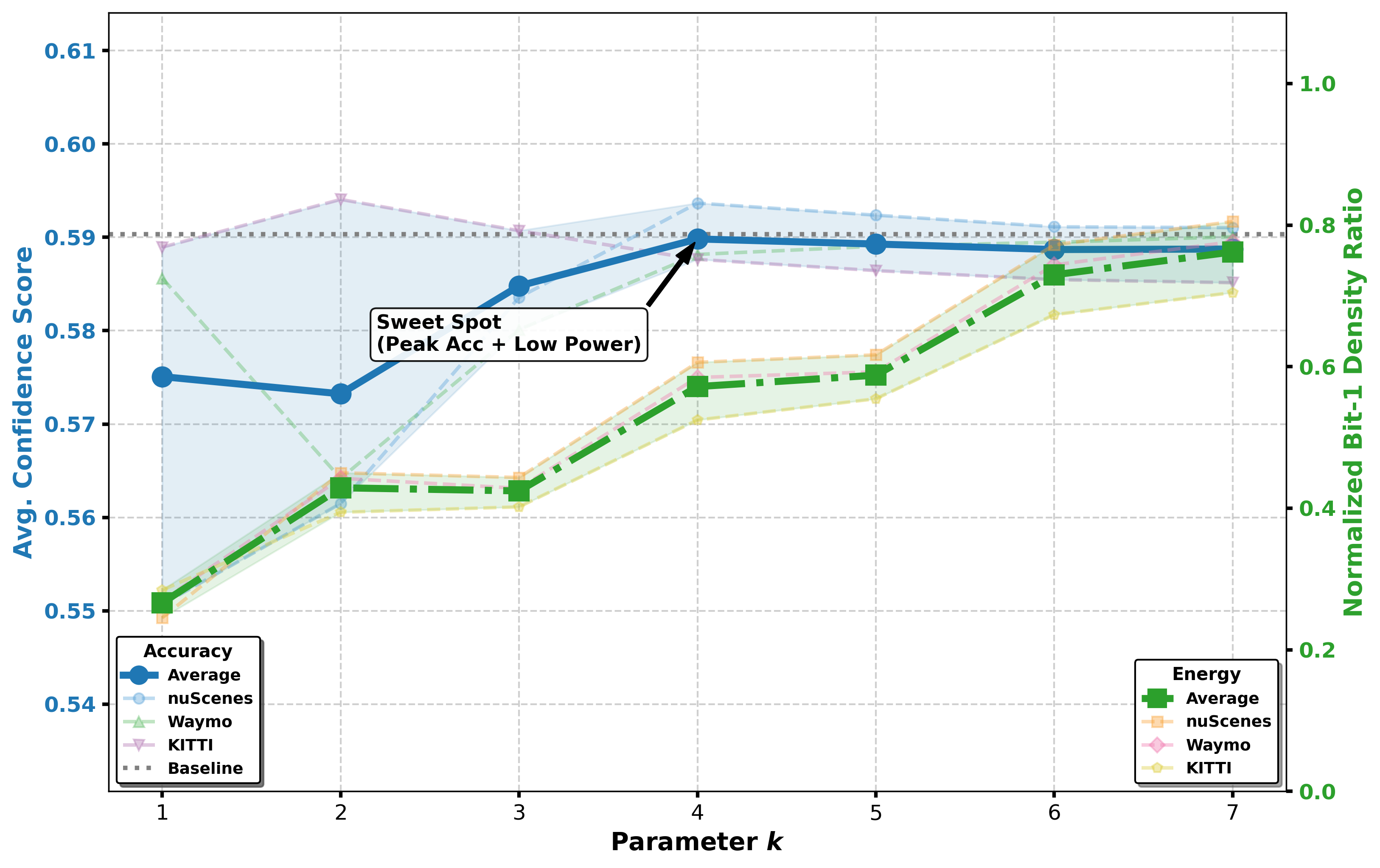}
	\caption{\textbf{Design Space Exploration: Perception vs. Energy Trade-off varying bit-width $k$.}
		The plot sweeps the retained parameter $k$ from 1 to 7.
		The \textcolor{blue}{blue line} (Accuracy/Confidence) saturates around $k=4$, while the \textcolor{green}{green line} (Energy Cost) continues to rise linearly.
		This identifies \textbf{$k=4$} as the algorithmic "Sweet Spot," providing maximum perceptual gain for the minimum necessary energy expenditure.}
	\label{fig:sensitivity}
	\vspace{-20pt}
\end{figure}


\begin{figure*}[th]
	\centering
	\includegraphics[width=\linewidth]{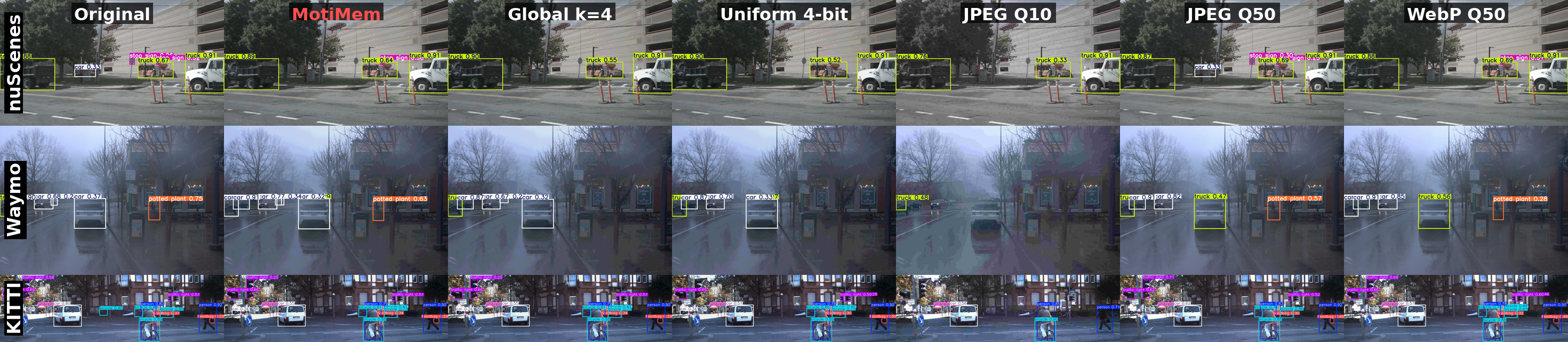}
	\caption{Qualitative comparison of detection results across three autonomous driving datasets (nuScenes, Waymo, KITTI) under different encoding methods. All images are processed by YOLO26m detector. MotiMem preserves detection accuracy comparable to the Original while significantly reducing energy consumption (43\% reduction). JPEG Q10 exhibits severe compression artifacts leading to missed detections. Standard codecs (JPEG Q50, WebP Q50) maintain visual quality but offer no energy benefit.}
	\label{fig:qualitative_results}
	\vspace{-20pt}
\end{figure*}

\begin{table*}[th]
	\centering
	\caption{Comprehensive comparison of encoding methods across three autonomous driving datasets. All metrics are averaged over 16 detection models. MotiMem achieves the \textbf{optimal trade-off}: highest mAP retention among energy-efficient methods while consuming only 57\% of the original energy.}
	\label{tab:full_results}
	\renewcommand{\arraystretch}{0.5}
	\setlength{\tabcolsep}{3.5pt}
	\scriptsize
	\begin{tabular}{l|ccccc|ccccc|ccccc|ccccc}
		\toprule
		\multirow{2}{*}{\textbf{Method}} & \multicolumn{5}{c|}{\textbf{nuScenes}} & \multicolumn{5}{c|}{\textbf{Waymo}} & \multicolumn{5}{c|}{\textbf{KITTI}} & \multicolumn{5}{c}{\textbf{Average}}                                                                                                                                                                                                                             \\
		                                 & mAP                                    & Eng.                                & SSIM                                & PSNR                                 & LPIPS & mAP  & Eng. & SSIM & PSNR     & LPIPS & mAP  & Eng. & SSIM & PSNR     & LPIPS & mAP                            & Eng.                  & SSIM                   & PSNR                   & LPIPS                  \\
		\midrule
		\multicolumn{21}{l}{\textit{Lossless Baselines}}                                                                                                                                                                                                                                                                                                                                                                         \\
		Original (8-bit)                 & 1.00                                   & 1.00                                & 1.00                                & $\infty$                             & .000  & 1.00 & 1.00 & 1.00 & $\infty$ & .000  & 1.00 & 1.00 & 1.00 & $\infty$ & .000  & 1.00                           & 1.00                  & 1.00                   & $\infty$               & .000                   \\
		JPEG2000                         & 1.00                                   & 1.02                                & 1.00                                & $\infty$                             & .000  & 1.00 & .94  & 1.00 & $\infty$ & .000  & 1.00 & 1.02 & 1.00 & $\infty$ & .000  & 1.00                           & 1.00                  & 1.00                   & $\infty$               & .000                   \\
		\midrule
		\multicolumn{21}{l}{\textit{MotiMem Variants (Ours)}}                                                                                                                                                                                                                                                                                                                                                                    \\
		\rowcolor{green!15}
		\textbf{MotiMem}$^\star$         & .91                                    & .63                                 & .92                                 & 30.6                                 & .215  & .92  & .51  & .93  & 31.8     & .198  & .95  & .59  & .91  & 29.4     & .228  & \textcolor{blue}{\textbf{.93}} & \textbf{.57}          & .92                    & 30.6                   & .214                   \\
		Global k=4                       & .90                                    & .59                                 & .92                                 & 30.2                                 & .230  & .90  & .49  & .93  & 31.4     & .212  & .93  & .49  & .91  & 29.0     & .243  & .91                            & .52                   & .92                    & 30.2                   & .228                   \\
		Uniform 4-bit                    & .90                                    & .48                                 & .91                                 & 29.5                                 & .231  & .90  & .49  & .92  & 30.6     & .214  & .93  & .44  & .90  & 28.3     & .245  & .91                            & \textcolor{blue}{.47} & .91                    & \textcolor{red}{29.5}  & .230                   \\
		\midrule
		\multicolumn{21}{l}{\textit{Standard Codecs}}                                                                                                                                                                                                                                                                                                                                                                            \\
		JPEG Q10                         & .74                                    & .73                                 & .85                                 & 31.7                                 & .389  & .70  & .70  & .86  & 32.9     & .372  & .74  & .88  & .84  & 30.5     & .401  & \textcolor{red}{.73}           & .77                   & \textcolor{red}{.85}   & 31.7                   & \textcolor{red}{.387}  \\
		JPEG Q50                         & .91                                    & .95                                 & .97                                 & 39.7                                 & .088  & .92  & .86  & .97  & 41.2     & .076  & .90  & .99  & .96  & 38.4     & .098  & .91                            & .93                   & .97                    & 39.8                   & .087                   \\
		JPEG Q100                        & .98                                    & 1.23                                & 1.00                                & 54.0                                 & .003  & .98  & 1.11 & 1.00 & 55.8     & .002  & .96  & 1.03 & 1.00 & 52.6     & .004  & .97                            & \textcolor{red}{1.12} & \textcolor{blue}{1.00} & \textcolor{blue}{54.1} & \textcolor{blue}{.003} \\
		WebP Q10                         & .80                                    & 1.04                                & .88                                 & 33.8                                 & .327  & .83  & .95  & .89  & 35.1     & .308  & .83  & 1.03 & .87  & 32.6     & .342  & .82                            & 1.01                  & .88                    & 33.8                   & .326                   \\
		WebP Q50                         & .88                                    & 1.03                                & .94                                 & 37.4                                 & .182  & .90  & .95  & .95  & 38.8     & .165  & .90  & 1.03 & .93  & 36.1     & .196  & .89                            & 1.00                  & .94                    & 37.4                   & .181                   \\
		WebP Q100                        & 1.00                                   & 1.05                                & .99                                 & 47.1                                 & .017  & 1.00 & 1.01 & 1.00 & 48.6     & .014  & 1.00 & 1.19 & .99  & 45.8     & .021  & 1.00                           & 1.08                  & .99                    & 47.2                   & .017                   \\
		\bottomrule
	\end{tabular}
	\vspace{1mm}

	\footnotesize{$^\star$\textbf{Pareto optimal}: MotiMem is the only method achieving both high mAP ($>$0.92) and low energy ($<$0.6). \\
		In Average column: \textcolor{blue}{blue} = best, \textcolor{red}{red} = worst among lossy methods. Eng. = Normalized Bit-1 Density ($\downarrow$).}
	\vspace{-15pt}
\end{table*}

\begin{table}[h]
	\centering
	\caption{Relative mAP retention (\%) across 16 detection models. Results averaged over 3 datasets. MotiMem achieves highest retention on all models.}
	\label{tab:model_comparison}
	\renewcommand{\arraystretch}{0.35}
	\small
	\begin{tabular*}{\columnwidth}{@{\extracolsep{\fill}}l|ccccc@{}}
		\toprule
		\textbf{Model} & \textbf{MotiMem} & \textbf{Glo. k=4} & \textbf{Uni. 4b} & \textbf{JPEG} & \textbf{WebP} \\
		\midrule
		YOLOv5n & \textbf{91.9} & 90.4 & 90.5 & 91.7 & 89.8 \\
		YOLOv8n & \textbf{92.0} & 90.6 & 90.6 & 91.6 & 89.7 \\
		YOLOv8s & \textbf{93.0} & 91.7 & 91.7 & 92.7 & 90.7 \\
		YOLOv8m & \textbf{93.5} & 91.9 & 91.9 & 91.8 & 90.0 \\
		YOLOv8l & \textbf{93.5} & 92.1 & 92.2 & 91.3 & 90.1 \\
		YOLOv8x & \textbf{93.7} & 92.3 & 92.3 & 91.3 & 89.9 \\
		YOLOv9c & \textbf{93.7} & 92.2 & 92.2 & 91.3 & 90.3 \\
		YOLOv10n & \textbf{92.6} & 91.2 & 91.2 & 92.8 & 90.8 \\
		YOLO11n & \textbf{92.3} & 90.8 & 90.8 & 91.9 & 89.7 \\
		YOLO11s & \textbf{92.6} & 90.8 & 90.8 & 90.9 & 89.2 \\
		YOLO11m & \textbf{93.7} & 91.8 & 91.9 & 91.5 & 90.4 \\
		YOLO11l & \textbf{94.0} & 92.4 & 92.4 & 91.8 & 90.6 \\
		YOLO11x & \textbf{94.2} & 92.7 & 92.6 & 91.6 & 90.6 \\
		YOLO26m & \textbf{92.8} & 90.8 & 90.9 & 90.5 & 88.9 \\
		RT-DETR-l & \textbf{89.2} & 87.7 & 87.7 & 87.3 & 84.7 \\
		RT-DETR-x & \textbf{89.6} & 88.0 & 88.0 & 87.8 & 85.2 \\
		\midrule
		\rowcolor{gray!15}
		\textbf{Average} & \textbf{92.6} & 91.1 & 91.1 & 91.1 & 89.4 \\
		\bottomrule
	\end{tabular*}
	\vspace{0.5mm}

	\footnotesize{Glo.: Global, Uni. 4b: Uniform 4-bit, JPEG/WebP: Q50.}
	\vspace{-20pt}
\end{table}

\vspace{-5pt}
\subsection{Sensitivity Analysis and Design Space Exploration}
\label{sec:sensitivity}

To determine the optimal operating point for the MotiMem interface, we analyze the sensitivity of both perception robustness and energy cost to the retained bit-width parameter $k$. We perform a sweep from $k=1$ to $k=7$ across the combined dataset. Fig.~\ref{fig:sensitivity} illustrates these opposing trends.

\vspace{-2.5pt}
\paragraph{The Energy Trend (Green Line)}
The green dashed line in Fig.~\ref{fig:sensitivity} illustrates the energy cost. As we retain more bits ($k > 4$), the Normalized Bit-1 Density rises strictly monotonically.
Specifically, increasing $k$ from 4 to 7 raises the density ratio from $\approx 0.58$ to $\approx 0.78$ (a $\approx 35\%$ increase).
This confirms that the lower bits in natural images are dominated by high-entropy noise. Including them contributes significantly to bus switching activity (cost) without providing meaningful structural information, as evidenced by the plateauing accuracy (Blue Line).
\vspace{-4pt}
\paragraph{The Perception Trend (Blue Line)}
The blue solid line represents the average confidence score of the detection models, which serves as a proxy for feature integrity.
We observe a law of diminishing returns:
The accuracy exhibits two distinct phases: a \textbf{Growth Phase} ($k=1 \to 3$), where accuracy improves drastically as the top MSBs define object boundaries and shapes, and a \textbf{Saturation Phase} ($k \ge 4$), where accuracy plateaus, matching baseline performance. Increasing $k$ beyond 4 offers negligible perceptual gain.
\vspace{-1pt}

\paragraph{The "Sweet Spot" Decision}
Based on this intersection, we identify $k=4$ as the Pareto-optimal elbow point. It effectively filters out the high-frequency "noise" (lower 4 bits) that consumes energy but aids little in detection, while preserving the high-fidelity "signal" (upper 4 bits + inversion) required by the neural networks. Consequently, all subsequent comparisons (Pareto analysis in Fig.~\ref{fig:pareto_efficiency}) use $k=4$ as the default configuration for MotiMem.

\subsection{Main Results}
Table~\ref{tab:full_results} presents the comprehensive comparison across 16 models and three datasets. We analyze the results focusing on the critical correlation between bit-statistics and hardware energy.

\paragraph{Energy-Accuracy Trade-off (The Physics of Saving)}
MotiMem establishes the optimal Pareto operating point. As detailed in Table~\ref{tab:full_results}, our method reduces the \textbf{Normalized Bit-1 Density} to \textbf{0.57}.
Since the dynamic power in memory buses is governed by the switching activity factor ($\alpha$), and bit-1 density serves as a linear proxy for $\alpha$, this density reduction \textbf{directly translates to a 43\% saving in dynamic energy consumption}.
In stark contrast, standard codecs fail to exploit this physical correlation: JPEG Q50 consumes significantly more energy (Density 0.93) for comparable accuracy, while JPEG Q100 even \textit{increases} energy consumption (Density 1.12) due to high-entropy coding overhead, offering negative energy benefits despite volume compression.

\paragraph{Impact of Semantic Awareness (Ablations)}
The comparison with ablations further validates our design. Although the \textit{Uniform 4-bit} variant achieves the lowest density (0.47), its lack of adaptive inversion leads to poor signal fidelity (PSNR 29.5 dB).
Crucially, the semantically blind \textit{Global k=4} baseline drops detection accuracy to 0.91 mAP.
MotiMem recovers this loss (+0.02 mAP) by protecting the RoI with high-fidelity bits.
Notably, MotiMem exhibits a lower PSNR (30.6 dB) than WebP Q50 (37.4 dB) yet achieves higher detection accuracy (0.93 vs. 0.89 mAP), proving that \textbf{optimizing the bit-density of semantic regions is more effective for energy-constrained perception than maximizing global pixel fidelity (PSNR).}

\paragraph{Robustness Across Models and Datasets}
MotiMem demonstrates consistent generalization. Across datasets, it maintains high accuracy on KITTI (0.95), Waymo (0.92), and nuScenes (0.91).
Regarding model sensitivity, larger backbones exhibit higher resilience to our density-optimized encoding: \textbf{YOLO11x} achieves an impressive \textbf{94.2\%} mAP retention. Even for sensitivity-prone Transformer models (RT-DETR), MotiMem maintains $\approx$89.6\% retention, consistently outperforming standard codecs which suffer from block-artifact induced degradation.

\vspace{-2.5pt}
\subsection{Discussion and Limitations}
\label{sec:discussion}

The disconnect between PSNR and mAP in Table I (MotiMem: 30.6 dB/0.93 mAP vs. WebP: 37.4 dB/0.89 mAP) confirms a fundamental insight: for neural perception, background fidelity is often energy waste. By strictly prioritizing Motion-Guided RoIs, we demonstrate that semantic-aware bit allocation lowers the energy barrier for edge devices more effectively than statistical compression.
However, we acknowledge two limitations: (1) Our 43\% energy savings are based on analytical bit-density proxies ($P \propto \alpha C V^2 f$) and require future validation on physical silicon; (2) The reliance on $T{-}1$ feedback prevents optimization during ``Cold Start'' or abrupt object entries. The inflation margin $\delta$ is set to cover typical inter-frame motion and localization errors, providing tolerance for newly appearing objects near existing detections, but a global-precision fallback remains necessary for entirely novel scene regions.
\vspace{-2.5pt}
\section{Conclusion}
\label{sec:conclusion}

We proposed \textbf{MotiMem}, a hardware-software co-designed interface that decouples semantic fidelity from background noise. By leveraging \textbf{Kinematic Priors} and \textbf{Hybrid Sparsity-Aware Coding}, MotiMem reduces the memory-interface dynamic energy proxy by $\approx 43\%$ while retaining $\approx 93\%$ of the detection mAP ($k=4$), significantly outperforming standard codecs.
Future work will focus on implementing the MotiMem logic on an \textbf{FPGA prototype} to validate physical power savings and extending the framework to \textbf{3D LiDAR point cloud perception}, where data sparsity can be further exploited for system-level efficiency.



\vspace{-2.5pt}

\bibliographystyle{IEEEtran}
\bibliography{ref}

\end{document}